RESEARCH ARTICLE

# Exploit fully automatic low-level segmented PET data for training high-level deep learning algorithms for the corresponding CT data

Christina Gsaxner[1,2,3], Peter M. Roth[1], Jürgen Wallner[2,3], Jan Egger[1,2,3]*

**1** Institute for Computer Graphics and Vision, Faculty of Computer Science and Biomedical Engineering, Graz University of Technology, Graz, Austria, **2** Computer Algorithms for Medicine Laboratory, Graz, Austria, **3** Department of Oral & Maxillofacial Surgery, Medical University of Graz, Auenbruggerplatz, Styria, Austria

* egger@tugraz.at, j.wallner@medunigraz.at

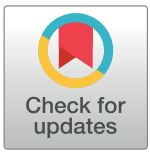

**Data Availability Statement:** All relevant data are available from https://wiki.cancerimagingarchive.net/display/Public/RIDER+Lung+PET-CT. The relevant code has been deposited in GitHub at https://github.com/cgsaxner/UB_Segmentation.

**Funding:** This work received funding from the Austrian Science Fund (FWF) KLI 678-B31: "enFaced: Virtual and Augmented Reality Training and Navigation Module for 3D-Printed Facial Defect Reconstructions" (PIs: Jürgen Wallner and Jan Egger) and the TU Graz Lead Project (Mechanics,

## Abstract

We present an approach for fully automatic urinary bladder segmentation in CT images with artificial neural networks in this study. Automatic medical image analysis has become an invaluable tool in the different treatment stages of diseases. Especially medical image segmentation plays a vital role, since segmentation is often the initial step in an image analysis pipeline. Since deep neural networks have made a large impact on the field of image processing in the past years, we use two different deep learning architectures to segment the urinary bladder. Both of these architectures are based on pre-trained classification networks that are adapted to perform semantic segmentation. Since deep neural networks require a large amount of training data, specifically images and corresponding ground truth labels, we furthermore propose a method to generate such a suitable training data set from Positron Emission Tomography/Computed Tomography image data. This is done by applying thresholding to the Positron Emission Tomography data for obtaining a ground truth and by utilizing data augmentation to enlarge the dataset. In this study, we discuss the influence of data augmentation on the segmentation results, and compare and evaluate the proposed architectures in terms of qualitative and quantitative segmentation performance. The results presented in this study allow concluding that deep neural networks can be considered a promising approach to segment the urinary bladder in CT images.

## Introduction

Since imaging modalities like computed tomography (CT) are widely used in diagnostics, clinical studies and, treatment planning and evaluation, automatic algorithms for image analysis have become an invaluable tool in medicine. Image segmentation algorithms are of special interest, since segmentation plays a vital role in various medical applications [1]. Typically, segmentation is the first step in a medical image analysis pipeline and therefore incorrect segmentation affects any subsequent steps heavily. However, automatic medical image segmentation is known to be one of the more complex problems in image analysis [2]. Therefore, to this day delineation is often done manually or semi-manually, especially in regions with limited contrast and for organs or tissues

Modeling and Simulation of Aortic Dissection). Moreover, this work was supported by CAMed (COMET K-Project 871132) which is funded by the Austrian Federal Ministry of Transport, Innovation and Technology (BMVIT) and the Austrian Federal Ministry for Digital and Economic Affairs (BMDW) and the Styrian Business Promotion Agency (SFG).

with large variations in geometry. This is a tedious task, since it is time consuming and requires a lot of empirical knowledge. Furthermore, the process of manual segmentation is prone to errors and since it is highly operator dependent, not reproducible, which emphasizes the need for accurate, automatic algorithms. One up-to-date method for automatic image segmentation is the usage of deep neural networks. In the past years, deep learning approaches have made a large impact in the field of image processing and analysis in general, outperforming the state of the art in many visual recognition tasks, e.g. in [3]. Artificial neural networks have also been applied successfully to medical image processing tasks such as segmentation.

For segmentation of the urinary bladder, there are currently two main applications. In clinical practice, it is used in radiation treatment planning e.g. treatment of prostate cancer. Prostate cancer is the most common cancer in northern and western European males, and the second most common cancer in men worldwide [4], [5]. External beam radiation therapy plays a critical part in the treatment of both localized and advanced cases. Since the radiation beam also damages normal cells, it is desirable to exclude as much healthy tissue from radiation as possible. Therefore, critical organs in the target region must be segmented beforehand to plan the radiation beam accordingly. The urinary bladder is, besides the rectum, the main organ that should be protected from radiation toxicities in the treatment of prostate cancer. However, delineation is mostly done manually, which presents a large workload for radiologists and comes with uncertainties. Hence, accurate, reliable and fully automatic segmentation is highly desirable. Furthermore, segmentation of the urinary bladder is a key step in computer-aided detection of urinary track abnormalities, such as bladder cancer. Cancer in the urinary bladder is the ninth most common cancer in the world [6]. Medical imaging modalities, especially computed tomography, play a big role in detecting and staging bladder cancer. However, the interpretation of those medical images is tedious and time consuming, as each individual slice has to be evaluated for lesions. Furthermore, the process leads to a substantial variability between radiologists in the detection of cancer, and there is also the chance of missing small lesions due to the large workload. Computer aided detection (CAD) might aid radiologists in finding lesions in the bladder. The first step in a CAD system is to define a search region for further detection, specifically to segment the urinary bladder. By excluding non-bladder structures for the search process, the possibility of false positive detections is decreased. Therefore, accurate bladder segmentation is a critical component in the computer aided detection of bladder cancer. In the treatment of urinary bladder cancer, radiation therapy, again, plays a critical part. However, since the urinary bladder is an organ which shows significant variations in size and position between patients and even within patients between individual therapy sessions, which limits the allowed radiation dose and results in large amounts of healthy tissue receiving the same radiation dose. Therefore, a technique called adaptive radiotherapy is used to re-optimize the plan during treatment to account for deformations of the target. Usually, a cone beam CT is taken before every treatment to select the optimal plan for the day. For quick and accurate plan selection, automatic segmentation of the urinary bladder in these CT scans is desirable. Additional applications include the measurement of parameters such as bladder wall thickness or bladder volume, which are critical indexes for many bladder-related conditions [7]. For example, bladder wall thickness can be a useful parameter in the evaluation of benign prostatic hyperplasia, a non-cancerous enlargement of the prostate and has been shown to be a useful predictor for bladder outlet obstruction and detrusor overactivity. Furthermore, focal bladder wall thickening is a sign for bladder cancer. Measuring bladder volume can be useful to look for urinary retention. These procedures would highly benefit from a fully automatic segmentation algorithm by reducing workload for physicians.

Therefore, we propose an approach for fully automatic urinary bladder segmentation in CT images using deep learning convolutional neural networks. Training such networks requires a large amount of labelled training data, which still presents a major bottleneck in the medical imaging field due to the highly sensitive nature of medical data and the large workload of manual segmentation. Therefore, we further propose a novel course of action to automatically generate the ground-truth labels from PET acquisitions to train neural networks for semantic segmentation.

## Material and methods

### Datasets and preprocessing

The implemented methods were trained and tested on the Reference Image Database to Evaluate Therapy Response (RIDER) [8]. RIDER is a collection of Computed Tomography, Magnetic Resonance Imaging (MRI) and PET/CT data. The RIDER PET/CT dataset provides serial patient studies (without meta-data) as well as data from multi-vendor and multi-parameter calibration phantoms. It consists of de-identified Digital Imaging and Communications in Medicine (DICOM) serial PET/CT lung cancer patient data and provides serial scans of 28 lung cancer patients (a total of 65 scans), as well as data from studies with a long half-life calibration phantom. As radiotracer, fluorine-18-labelled fluorodeoxyglucose was used in all PET scans. The public database can be downloaded from the National Cancer Imaging Archive (NCIA) at [9].

Unfortunately, some datasets exhibit a large amount of artifacts in the region of interest. This is due to the bladders position within the pelvic bone. The high attenuation of the bone can lead to a larger amount of noise along the direction of greatest attenuation and furthermore to beam hardening and scatter. If one wants to image the bladder specifically, these artifacts can be reduced by using iterative reconstruction or special noise reduction algorithms. We removed the datasets with the largest amount of artifacts from the database. However, we also saw these disturbances as a chance for our neural network to learn invariance to them. After removing patient data with low contrast and high noise from the RIDER PET/CT database, a total of 29 patient datasets were obtained. The CT datasets offer between 148 and 358 transversal slices, yielding a total of 8754 CT scans. Since these scans cover the whole torso the urinary bladder is only visible on a fractional amount of the images, with an average of 25 transversal slices per dataset covering the bladder. It would not be sensible to train our deep network with such a large amount of negative training examples, therefore, a total amount of 845 CT image slices around the urinary bladder were extracted from the whole dataset.

### Generation of image data

The number of images obtained from the RIDER PET/CT database is probably not sufficient to train a deep neural network. It also has to be considered that not all data can be used for training, since testing data must also be taken from the same dataset. Furthermore, a reference standard in the form of segmented CT images is not available. The first problem, the small number of training samples, is solved by using data augmentation. The process of data augmentation has become a "must-do" when training deep neural networks. By applying transformations to the existing images, we can provide our network with additional images that are similar to the existing ones, but not exactly the same. As long as the transformations are meaningful within the context of the task, the network can learn invariance to the applied changes and should therefore be able to generalize better to new, unknown data. Of course, the amount of images that can be created by augmentation is limited. If the applied changes are too large, the network will not be able to extract information meaningful for the task at hand anymore. If the applied changes are too small, there won't be any additional information for the network [10]. To enlarge our dataset, rotation and scaling are applied to CT images as well as the masks generated from the corresponding PET data. Furthermore, zero-mean Gaussian noise was added to CT images.

The generation of segmentations of the urinary bladder as a ground truth for training a deep neural network is performed by using combined positron emission tomography-computed tomography scans. The most commonly used radiotracer, fluorine-18-labelled fluorodeoxyglucose ($^{18}$F-FDG), accumulates in the urinary bladder, therefore, the bladder always shows up in these PET scans. Contrary to CT, PET images exhibit high contrast and are therefore comparably easy to segment automatically, which is highlighted in Fig 1. We automatically segment PET images using a simple thresholding approach to generate binary masks of the urinary bladder that match CT data acquired from the same patients at the same time. The necessary steps for data generation were implemented in the modular medical imaging framework MeVisLab (www.mevislab.de).

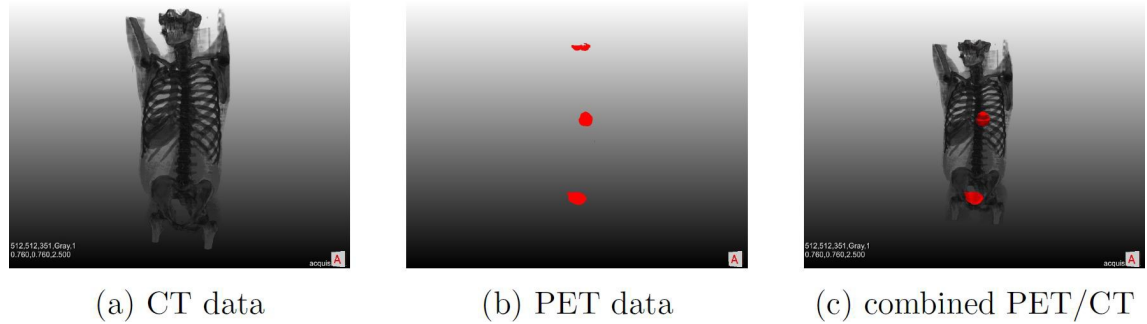

(a) CT data      (b) PET data      (c) combined PET/CT

**Fig 1. 3D image data obtained from CT, PET and combined PET/CT of the torso and part of the head.** While CT data in (a) shows important anatomical structures, the contrast for soft tissue, in example in the abdominal region, is poor. PET data in (b) only shows metabolically active regions, without providing anatomical context, making it impossible to accurately localize lesions. In the co-registered PET/CT scan in (c), it is possible to properly assign active regions anatomically. The urinary bladder, a lesion in the left lung and parts of the brain are highlighted via the PET data.

The available 29 patient datasets containing a total of 845 image slices showing the urinary bladder were split into training and testing data. A standard for this split commonly found in literature is 80% training data, 20% testing data. Loosely following this guideline, 630 images were used for training and 215 images were reserved for testing, corresponding to 21 and 8 patient datasets, respectively. Next, the 21 patient datasets for training were processed with the proposed MeVisLab network to obtain individual, augmented CT slices as well as corresponding ground truth labels. For the 8 datasets reserved for testing, only a ground truth label was created and no augmentation was applied. To additionally be able to analyse the effect of data augmentation, a training dataset containing only the 630 un-augmented, original images and labels as well as a dataset with only transformed image data (without noise) was put together. The MeVisLab network and corresponding python code can be found on github [11].

The general network constructed in MeVisLab can be seen in Fig 2. The purpose of this network is to load process and visualize the given input data using modules already integrated in the MeVisLab framework, as well as a self-implemented macro module. Corresponding PET and CT image data is loaded into the module.These three-dimensional images are then fed into the *DataPreparation* macro module, which the centrepiece of the data generation step in this study. It calculates binary masks from the PET data, performs data augmentation as specified by the user and saves the created training data.

*The DataPreparation Macro Module*–The MeVisLab macro module consists of three files:

1. The internal network (*DataPreparation.mlab*). This file contains the internal network structure of the macro module. It contains a module for thresholding the PET portion of the data by calculating a fixed threshold for each dataset within the Python script. The threshold is defined as a fixed percentage, in this case 20%, of the maximal SUV within the dataset:

$$T = SUV_{max} \cdot 0.2$$

Every pixel above the threshold is considered foreground, all other pixels are labelled as background. For applying data augmentation to both PET and CT images, several affine transformations in 2D are applicable. For this macro module, rotation and scaling are enabled. Furthermore, zero-mean Gaussian noise can be added to CT images as an additional augmentation step. After processing, the resulting binary masks and corresponding CT images are saved. The internal network of the *DataPreparation* macro module can be seen in Fig 3.

2. The MeVisLab Definition Language (MDL) scripts file (*DataPreparation.script*). It consists of three main sections:

a. The interface section defines the inputs (CT and PET image data) and outputs (no outputs are used for this module) of data connections. Furthermore, the parameters fields of the implemented macro module are declared here. The fields declared in the interface section can either be independent script fields or they can be defined as aliases for internal fields of the internal network.

b. The commands section defines the scripting file containing the functions to be executed upon the activation of certain fields. Also, the commands for calling these functions are defined.

c. The window section can be used to create a panel for the macro module. The panel can be used to choose the parameters for data augmentation and for monitoring the progress of the file export.

3. The Python script files (*DataPreparation.py*). Here, functions and interactions between modules are implemented using Python scripting.

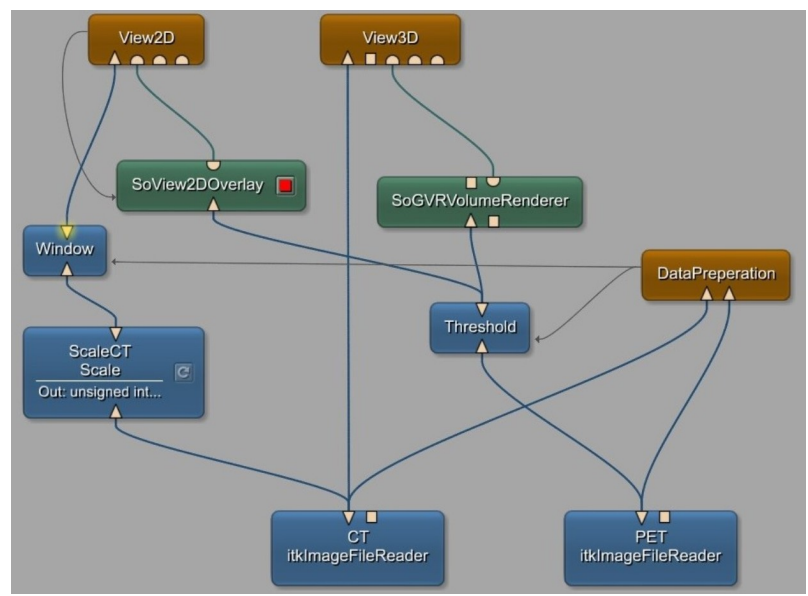

**Fig 2. General Network for loading, processing and visualizing PET and CT data, implemented in MeVisLab.**

https://doi.org/10.1371/journal.pone.0212550.g002

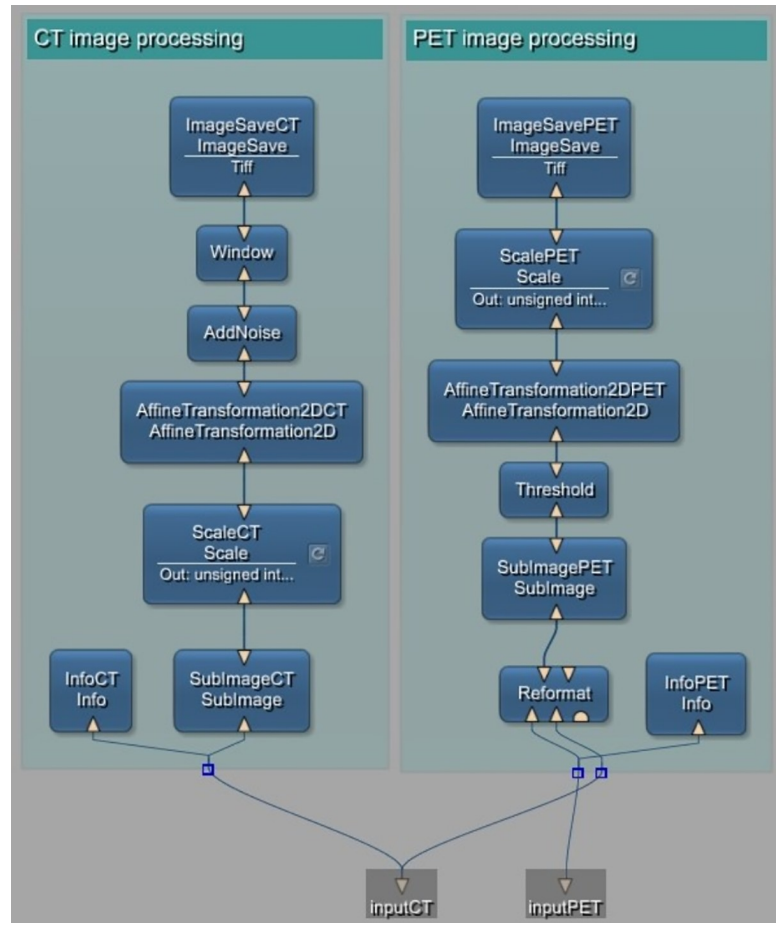

**Fig 3. The internal network of the DataPreparation macro module.**

## Network architectures

*Convolutional Networks for Semantic Segmentation (FCNs)*—Introduced by Long et al. in [12], they made a large impact on semantic segmentation with neural networks by providing end-to-end, pixel-to-pixel segmentation networks. The basic idea behind this approach is to transform established classification networks to fully convolutional networks, suitable for semantic segmentation. They achieved best results with the VGG-16 classification network [13]. For this, fully connected layers are transformed into convolution layers, which allow the network to output a spacial heatmap. However, these output maps are considerably downsampled from the input image because of pooling layers, making the output very coarse. To map these coarse outputs to dense predictions, upsampling layers, which resample the image to its original size, are proposed. For upsampling, not only the features from the last downsampling layer are used. Instead, so-called skip connections are introduced to use convolutional features from different layers in the network, which have different scales. Since shallower layers produce bigger feature maps where more spatial information is preserved, this helps capturing finer details from the original image. Upsampling layers are learned for each of these skip connections individually. The best performing architecture is FCN-8s, where a combination of VGG-16 features upsampled by a factor of two and features from the fourth pooling layer is combined with features obtained from the third pooling layer of VGG-16 and then upsampled by a factor of eight. This is illustrated in Fig 4.

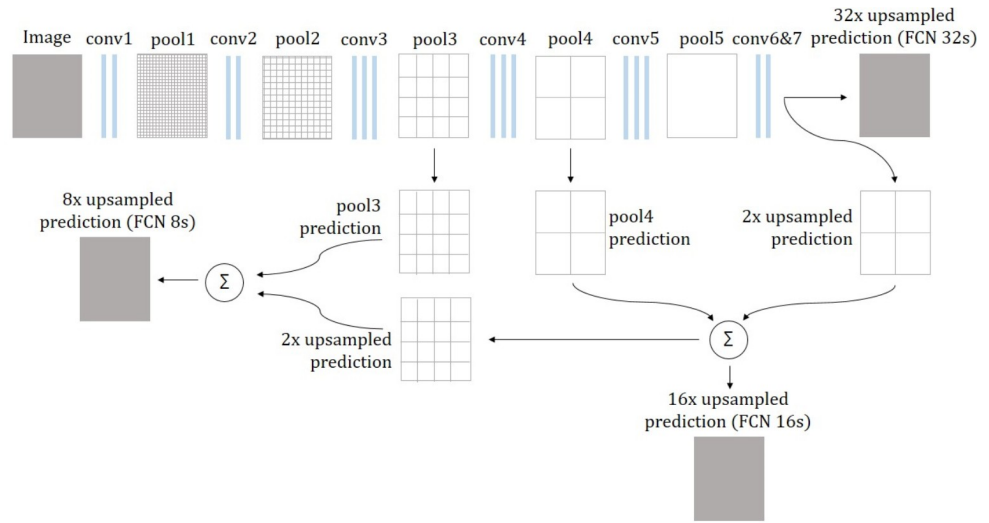

**Fig 4. FCN architectures.** The input image is downsampled by max pooling layers, getting coarser from layer to layer. In the FCN 32s, the output of the last pooling layer is upsampled by factor 32 in a single step. In FCN 16s, features from the last layer and pool4 layer are combined and then upsampled by factor 16. For FCN 8s, predictions from pool3 layer are included. Adapted from [12].

*Atrous Convolution for Semantic Segmentation*—Chen et al. [14] build on a different approach to deal with the problem of considerably downsampled feature maps resulting after a traditional classification networks. They utilize so-called atrous convolution, also called dilated convolution, which is convolution with upsampled filters. By replacing convolutional layers in classification networks such as VGG-16 or ResNet [15] with atrous convolution layers, the resolution of feature maps of any layer within a CNN can be controlled. Furthermore, the receptive field of filters can be enlarged without increasing the number of parameters. This convolution allows the construction of many layered networks without decreasing resolution, and since only non-zero values have to be accounted for in convolutions, the number of filter parameters does not increase. However, computing feature maps at the original image resolution is not very efficient, therefore, hybrid approaches are mostly used. Atrous convolution layers are applied in a way to downsample the original image by a factor of eight in total (compared to a factor of 32 in VGG Nets or ResNets), followed by bilinear interpolation to recover feature maps of the original image resolution. Compared to approaches using fractionally strided convolution for upsampling, no new parameters are learned within the network, which leads to faster training. Atrous convolution in 2-D is illustrated in Fig 5.

## Implementation

Algorithms for image segmentation using deep neural networks were implemented using TensorFlow 1.3 under Python 3.5. The code can be found on github [17]. It is divided up into several files:

- *tf_records.py*: Contains functions for creating and reading from the TensorFlow recommended file format TFRecords. Those functions are used to transform image data into a file format that is easy and fast to process in TensorFlow.

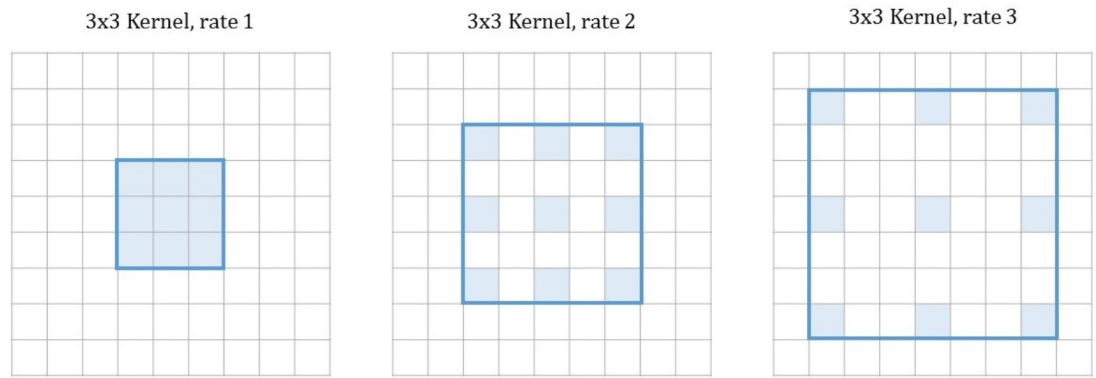

**Fig 5. Atrous convolution.** In all examples the kernel size is 3x3, but the rate differs. The rate defines by which factor the filter is dilated. Empty values are filled with zeroes. The larger the rate, the larger the receptive field of the filter becomes. For a rate of one, atrous convolution corresponds to standard convolution. Adapted from [16].

https://doi.org/10.1371/journal.pone.0212550.g005

- *make_tfrecords_dataset.py*: This script can be used to put together a TFRecords dataset from a directory of image files.

- *networks.py*: Includes the model definitions for the implemented image segmentation networks. These functions can be run in training or testing mode.

- *upsampling.py*: Contains tools for creating bilinear upsampling filters used for upsampling the predictions made by the networks using transposed convolution.

- *FCN_training.py* and *ResNet_training.py*: These scripts are used for training the deep neural network models defined in *networks.py*.

- *FCN_testing.py* and *ResNet_testing.py*: These scripts can be used for testing the previously trained deep neural networks.

- *metrics.py*: Provides metrics for evaluating the segmentation results by calculating similarity measures between network prediction and ground truth.

**Using pre-trained networks.** Neural networks for complex tasks like image segmentation need to be large and deep, resulting in many thousands of parameters. This means that training such networks requires huge datasets and a lot of computational power, and the training process might still require days or even weeks to complete. Therefore, pre-trained models were used for the segmentation task at hand. The TF-Slim API contains a set of standard model definitions implemented with TF-Slim as well as checkpoints for pre-trained parameters. An overview of the implemented models, corresponding code and links to the model checkpoints can be found at [18]. The models were trained on the ILSVRC-2012 dataset for image classification. The TF-slim model library contains pre-trained versions of VGG 16 and ResNet V2, which are adapted for segmentation tasks. The implementation of these models is based on the implementation found in [19]. The first network definition, FCN, uses a pre-trained version VGG 16, wich is adapted using skip connections and upsamling, resulting in the FCN 8s architecture as described by Long et al. in [12].Note that while upsampling filters could be defined as a learnable variable, they are kept fixed in this model, since Long et al. stated in their paper that learnable upsampling kernels didn't significantly improve the performance of the model, while making computation more expensive. The network architecture was illustrated using TensorBoard, which can be seen in Fig 6(A).

The second model definition, upsampled ResNet utilizes a pre-trained version of ResNet V2 with 152 layers and atrous convolution to perform image segmentation. The architecture was chosen in a way to achieve a downsampling of the input image by a factor of eight, which appeared to be a good trade-off between density of the computed feature maps and computational and memory expenditure. Again, a visualization of the network architecture was obtained using TensorBoard, as seen in Fig 6(B).

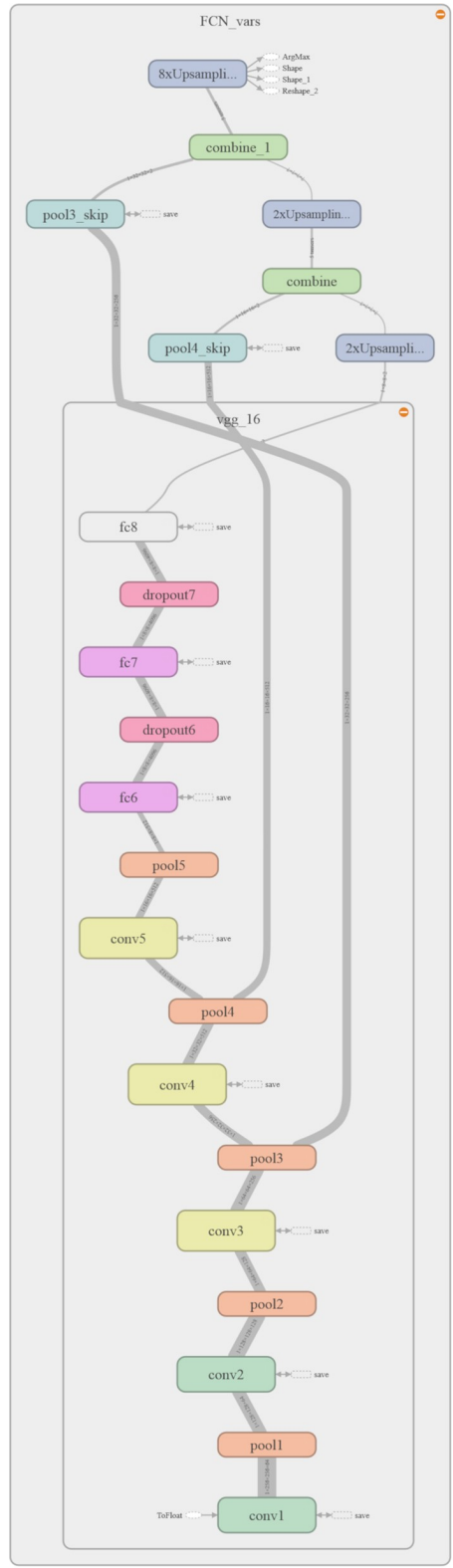

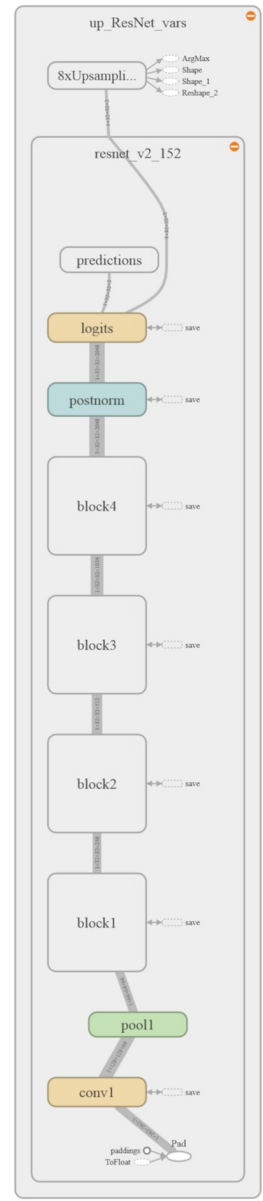

(a) FCN architecture      (b) upsampled ResNet architecture

**Fig 6. Network Graphs for FCN and upsampled ResNet visualized with TensorBoard.** Figure (a) shows the implemented FCN architecture, which is based on a pre-trained VGG 16 network and upsampling with skip connections. Figure (b) shows our upsampled ResNet architecture, which is based on a pre-trained ResNet V2 152 network. The layers of this network are condensed within 4 building blocks.

**Upsampling.** It was stated by Long et al. in [12] that upsampling can be performed by using transposed convolution. It is also often referred to as fractionally strided convolution or deconvolution. Convolution can be seen as sliding a convolution filter over an image and computing the dot product between filter and input in every step, which gives one element of the output image. Depending on the stride of the convolution filter, the resolution of the output image might be of lower resolution than the input. Fractionally strided convolution performs the opposite operation, going from a small resolution input to a bigger resolution output. One element in the input image defines the weights for the convolution filter, which is then copied to the output. Where filter regions overlap, the filter values are added. This operation is actually equivalent to the backward pass of a traditional convolution performed during error backpropagation. An illustration of this can be seen in Fig 7.

## Training

Training was performed using an ADAM optimizer over 34,020 iterations (corresponding to the size of the largest training data set) with a batch size of one. As a loss function, cross entropy loss was used. We trained on a server equipped with a NVIDIA Tesla K20Xm with 5 GB memory size. Testing was executed using a NVIDIA GeForce GTX 960 with 2 GB of memory. We trained our networks with images of original resolution (512x512) and downsampled images (256x256). Code for training the deep neural networks is found in the files *FCN_training.py* and *ResNet_training.py*. Both files follow the same pattern, however, since networks are trained from different checkpoint files, there are some differences between training a fully convolutional network and an upsampled ResNet network. While the FCN architecture requires a VGG 16 checkpoint to work, a ResNet V2 152 checkpoint is needed for the upsampled ResNet architecture. Segmentation is performed into two classes, background and foreground (urinary bladder), so the number of classes is chosen accordingly. Batch size is chosen as 1 as it was in the original architectures, furthermore, the small batch size avoided exceeding GPU memory. A batch size of 1 means that each image will be processed individually and no batch normalization is performed. Then, images and labels are loaded from a specified TFRecords training data file. As loss function, cross entropy is calculated between the logits of the model and the labels. As an optimizer, instead of the simple gradient descent algorithm, the more sophisticated optimizer following the Adam algorithm proposed in [20] with a learning rate of $10^{-4}$, as suggested in the paper, is used. While this optimizer requires more computations for each parameter in each training step, it usually converges more quickly without the need to fine tune the learning rate. The learning rate is chosen as a fixed value, since Adam optimizer performs learning rate decay internally

## Testing and evaluation

To evaluate the results achieved with the proposed neural networks, several metrics that are commonly applied for measuring similarity between the ground truth and the segmentation result when working with medical image data are calculated. True Positive Rate, True Negative Rate and Dice coefficient are metrics commonly found in literature. However, these metrics might be a poor measure for images with a lot of background and small object segments. Therefore, the Hausdorff distance is additionally measured. It is also a useful estimate when the boundary delineation of the segmentation is of special interest, as it is the case in this study. However, it should be noted that Hausdorff distance is very sensitive to outliers [21]. Functions for calculating evaluation metrics are found in the file *metrics.py*

The true positive rate (TPR), commonly referred to as sensitivity, measures the amount of positive pixels (foreground pixels) in the ground truth that are also identified as positives by the segmentation algorithm. The true negative rate (TNR), also called specificity, on the other hand, measures the portion of negative pixels (background pixels) in the ground truth segmentation that are correctly identified as such by the algorithm. The measures are defined as

$$TPR = \frac{TP}{TP + FN}$$

and

$$TNR = \frac{TN}{TN + FP}$$

where

- TP are the true positives, meaning pixels which are correctly classified to the foreground

- FN are false negatives, pixels that are incorrectly classified to the background

- TN are true negatives, pixels which are correctly assigned to background

- FP are false positives, meaning pixels that are incorrectly identified as foreground [22].

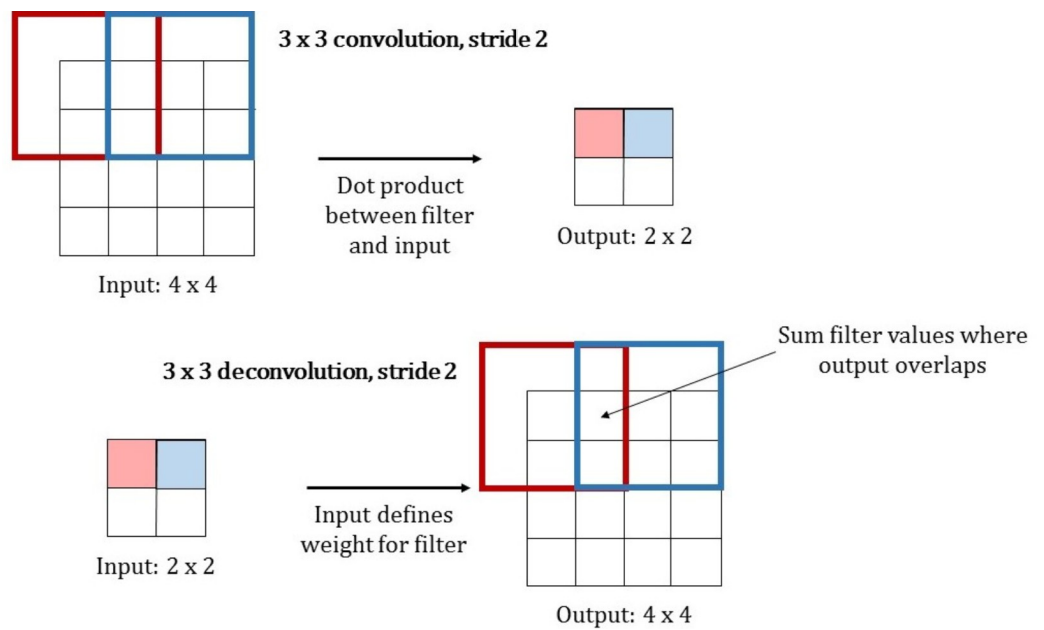

**Fig 7. Comparison between normal convolution and transposed convolution.** Both operations use a 3x3 kernel and a stride of two. Traditional convolution determines the output value as the dot product between filter and input, by moving the filter kernel for two pixels in every step, the input is downsampled by factor two. For transposed convolution, the input value determines the filter values that will be written to the output. Where filters overlap, the values are summed up. The stride defines the movement of the filter kernel in the output image, and therefore influences the factor of upsampling.

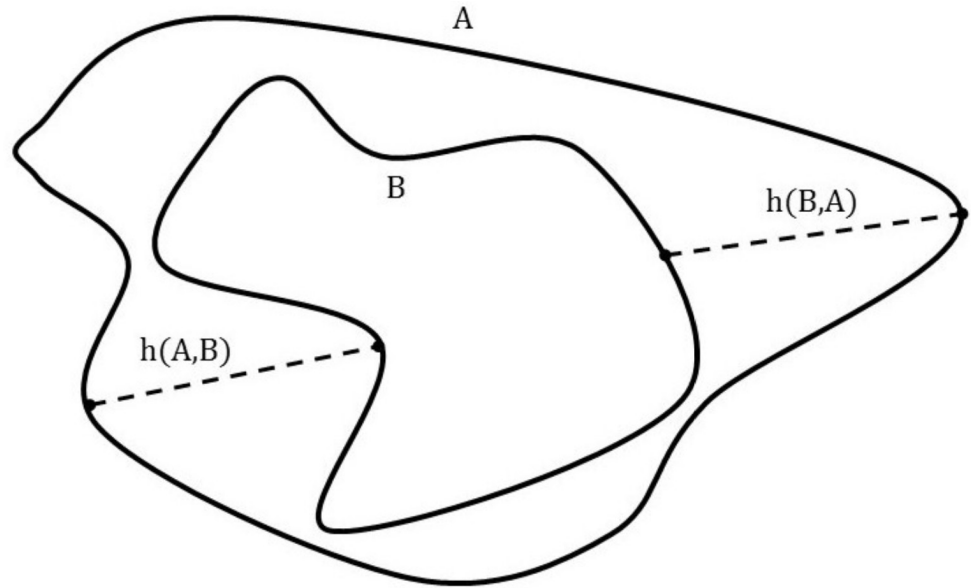

**Fig 8. Graphical interpretation of Hausdorff distance.** h(A,B) is the distance between the most distant point of point set A from the closest point of point set B. For h(B,A), it is opposite. HD is the maximum between h(A,B) and h(B,A). Adapted from [25].

https://doi.org/10.1371/journal.pone.0212550.g008

The Dice coefficient [23], also called Sorensen-Dice coefficient (DSC), is the most used metric for validating medical image segmentation. It is an overlap based metric. For a ground truth segmentation $S_g$ and a predicted segmentation $S_p$ the DICE can be calculated as

$$DSC = \frac{2|S_g \cap S_p|}{|S_g| + |S_p|}$$

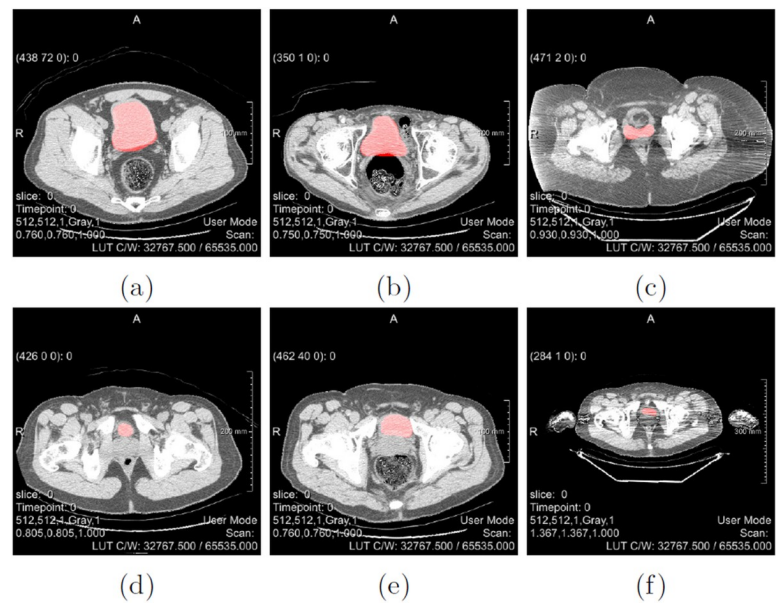

**Fig 9. Examples of overlays between CT data and generated ground truth labels.** The underlying CT images are shown in greyscale, while the ground truth labels obtained from PET segmentation are added in red.

https://doi.org/10.1371/journal.pone.0212550.g009

**Table 1. Parameters for data augmentation.** This table shows all parameters specifiable by the user, as well as their default, minimal and maximal values.

| Augmentation Type | | Parameter | Default | Minimum | Maximum |
|---|---|---|---|---|---|
| Rotation | | Maximal Rotation Angle | 45° | 0° | 180° |
| | | Number of Rotations per Slice | 4 | 1 | 10 |
| Scaling | | Maximal Scale Factor | 0.1 | 0.05 | 0.15 |
| | | Number of Scalings in x-Direction | 2 | 0 | 5 |
| | | Number of Scalings in y-Direction | 2 | 0 | 5 |
| Noise | | Number of Noisy Slices | 4 | 1 | 10 |
| | Uniform | Maximal Amplitude | 5 | 1 | 10 |
| | zero-mean Gaussian | Maximal Standard Deviation | 5 | 1 | 10 |
| | Salt and Pepper | Maximal Density | 0.2 | 0.05 | 0.5 |

where $2|S_g \cap S_p|$ is the intersection between ground truth segmentation and predicted segmentation. This intersection corresponds to the true positives TP. $|S_g|$ and $|S_p|$ denote the total amount of pixels classified to foreground in the ground truth and the prediction, respectively. The DSC takes values between 0 and 1, where 1 equals a perfect match.

The Hausdorff distance (HD) is a spatial distance based similarity measure, which means that the spatial position of pixels are taken into consideration. The Hausdorff distance between two point sets A and B is defined as

$$HD(A, B) = \max(h(A, B), h(B, A))$$

where h(A,B) is the directed Hausdorff distance. It describes the maximal distance of point set A to the closest point in point set B. It's mathematical definition is

$$h(A, B) + \max_{a \in A} \ \min_{b \in B} ||a - b||$$

where a and b are points of point set A and B respectively and $||\dots||$ is a norm, in example L2 norm to calculate Euclidian distance between the two points. A graphical representation of the Hausdorff distance and directed Hausdorff distance can be seen in Fig 8. The Euclidian distance can be calculated by [24].

$$||a - b||_2 = \sqrt{\sum_i (a_i - b_i)^2}$$

**Table 2. Segmentation evaluation results for images rescaled to 256x256.** This Table compares evaluation metrics for FCN and upsampled ResNet architectures trained using unaugmented training data, transformed training data (rotation, scaling) and fully augmented data (transformations and zero-mean Gaussian noise).

| Network Model | Training Data | mean TPR (%) | mean TNR (%) | mean DSC (%) | mean HD (pixel) |
|---|---|---|---|---|---|
| FCN | no augmentation | 82.7 | 99.9 | 77.6 | 6.9 |
| | **transformed images** | **85.0** | **99.9** | **80.4** | **6.1** |
| | fully augmented images | 79.2 | 99.9 | 77.6 | 6.7 |
| upsampled ResNet | no augmentation | 80.7 | 99.9 | 73.5 | 7.9 |
| | transformed images | 82.5 | 99.9 | 76.9 | 6.3 |
| | fully augmented images | 79.7 | 99.9 | 76.7 | 7.7 |

**Table 3. Segmentation evaluation results for images of resolution 512x512.** This Table compares evaluation metrics for FCN and upsampled ResNet architectures trained using unaugmented training data and transformed training data (rotation, scaling).

| Network Model | Training Data | mean TPR | mean TNR | mean DSC | mean HD |
|---|---|---|---|---|---|
| | | (%) | (%) | (%) | (pixel) |
| FCN | no augmentation | 80.9 | 99.9 | 77.6 | 13.3 |
| | **transformed images** | **83.1** | **99.9** | **81.9** | **11.9** |
| upsampled ResNet | no augmentation | 68.7 | 99.9 | 71.1 | 23.9 |
| | transformed images | 86.5 | 99.8 | 67.1 | 16.9 |

## Results

### Generation of training and testing data

To illustrate the agreement between the ground truth labels obtained from thresholding the PET data and corresponding CT images, overlays between CT images and generated labels were produced. Some examples of these overlays can be seen in Fig 9. By applying data augmentation with the default parameters specified in Table 1 to the original 630 training images, a total amount of 34,020 augmented images and labels were obtained. This equals a magnification factor of the original dataset by 54. A magnification factor of 27 was achieved by the transformation, specifically the combination of rotations and scaling of the input images. The amount of transformed data was then doubled by the addition of zero-mean Gaussian noise on each image slice. The remaining noise types included in the *DataPreparing* MeVisLab macro module have not yet been explored. By fully exploiting the maximal parameters defined in Table 1, a magnification factor of up to 1350 could be achieved, which would result in a total of 850,500 augmented image slices. However, since such large amounts of data are hard to handle with the available resources, such a large dataset was not created.

It can be observed that agreement between CT images and corresponding generated ground truth is generally good, but not perfect. Accuracy differs from dataset to dataset and even within individual slices. It can be observed that accuracy is worse in images were the urinary bladder only covers a small area surface of the image, like in image 10 (d) and (f). This is due to the nature of PET imaging, which has low spatial resolution and therefore, object boundaries might appear blurred. This is especially problematic when objects are small. Fig 9 also shows some of the unique challenges one is confronted with when automatically segmenting the urinary bladder in CT images. It can be noted that size and position of the urinary bladder is varying between patients. In some image slices, for example in Fig 9(C), the shape of the urinary bladder highly differs from its conventional, round form. Furthermore, low contrast between the bladder and surrounding soft tissue, as seen in Fig 9(B), poses a big difficulty. This especially occurs at the ambiguous bladder-prostate interface, as shown in Fig 9(E). It also becomes evident that not all CT data offers the same quality. In example, images 10 (c) and (f) show noticeable streak artefacts. Those artefacts are commonly found in CT scans and appear between dense objects like bone or metal due to beam hardening. Furthermore, since feature maps are significantly downsampled within our network architectures, images with a small area surface of the urinary bladder, as seen in 10 (f) might pose a problem, since small details could be lost as a result of downsampling.

### Image segmentation

Table 2 shows the results of segmentation evaluation for models trained with different training datasets at a resolution of 256x256. Table 3 presents the same metrics for segmentation results obtained from models with images at their original resolution. True positive rate, true negative rate, Dice coefficient, each in percent, and Hausdorff distance, in pixels, averaged over all 215 training datasets are listed.

These evaluation results show that the application of data augmentation in the form of scaling and rotation to the original dataset does improve segmentation performance significantly. For mean TPR and DSC an increase of 2.3% and 2.8%, respectively, was achieved with the FCN architecture and training and testing images with resolution of 256x256. For data with resolution 512x512 similar increases of 2.1% mean TPR and 4.3% mean DSC was achieved. With the upsampled ResNet architecture the corresponding enhancement was 1.8% and 3.5% with downsampled images or 17.8% TPR and a decrease of 4% DSC with original image resolution. Average Hausdorff distance was decreased by 0.8 pixels and 1.4 pixels in FCN models, as well as 1.6 pixels and 7 pixels using upsampled ResNet for the 256x256 and 512x512 datasets, respectively. The mean true negative rate exhibited very high values regardless of the used network model and shows no significant variations between different training data sets. Although transformation of training data did increase segmentation performance, it can be noted that segmentation results obtained from the network trained with the original dataset consisting of only 630 images and labels are also quite satisfactory for models trained with downsampled image resolution. This shows that when using pre-trained networks, one can achieve passable results with only a small amount of training data. From the evaluation scores achieved with models trained with fully augmented data in Table 2, it can be seen that the addition of noise to the transformed training dataset does not improve the performance of our proposed networks. In fact, all metrics show a worse performance of networks trained with artificially noisy training data compared to networks trained with only transformed training data. Using the FCN architecture, mean TPR even showed higher results when the network was only trained with 630 un-augmented training sets, with a TPR of 82.7% compared to 79.2% achieved with the fully augmented training set. The same is true for the upsampled ResNet architecture, although to a lesser extent, with an achieved TPR of 80.7% using un-augmented data compared to 79.7% using fully augmented data. One explanation for this could be that the applied Gaussian noise is not meaningful in the presented context. Therefore, the network learns spurious patterns that are not present in the training data. Another reason for the decrease in performance might be that the added noise is not strong enough. This results in the model seeing very similar images repeatedly, which might lead to overfitting. The model starts to fit too specific to the training set and loses its ability to generalize to the new examples found in the testing set. Since we didn't obtain satisfactory results with the noisy training data, network architectures were not trained with this data at its original resolution of 512x512.

Table 2 shows that in case of images rescaled to 256x256, best results can be achieved with the FCN architecture trained with images that are augmented with scaling and rotation. This network resulted in the highest true positive rate (85.0%) and Dice coefficient (80.4%) as well as the lowest Hausdorff distance (6.1 pixels). The true negative rate at 99.9% is the same for all tested models. The very high specificity indicates that our models are very accurate when it comes to correctly labelling background. However, this measure is highly dependent on segment size. Images with a lot of background, as it is the case in our examples, naturally show a higher TNR. Regardless of the used training data, the FCN architecture outperforms the upsampled ResNet architecture in all evaluation metrics. Inspecting the evaluation results using images at their original resolution of 512x512 in Table 3, again, the FCN architecture generally performs better in terms of our evaluation metrics. Especially Dice coefficient is notably higher at 81.9% for FCN than for ResNet at 67.1% for our best performing models. Also, the Hausdorff distance is shorter by 5 pixels, indicating that our upsampled ResNet architecture produces more outliers. Only in terms of true positive rate, the ResNet architecture achieved better results with a TPR of 86.5% compared to 83.1% for the FCN architecture.

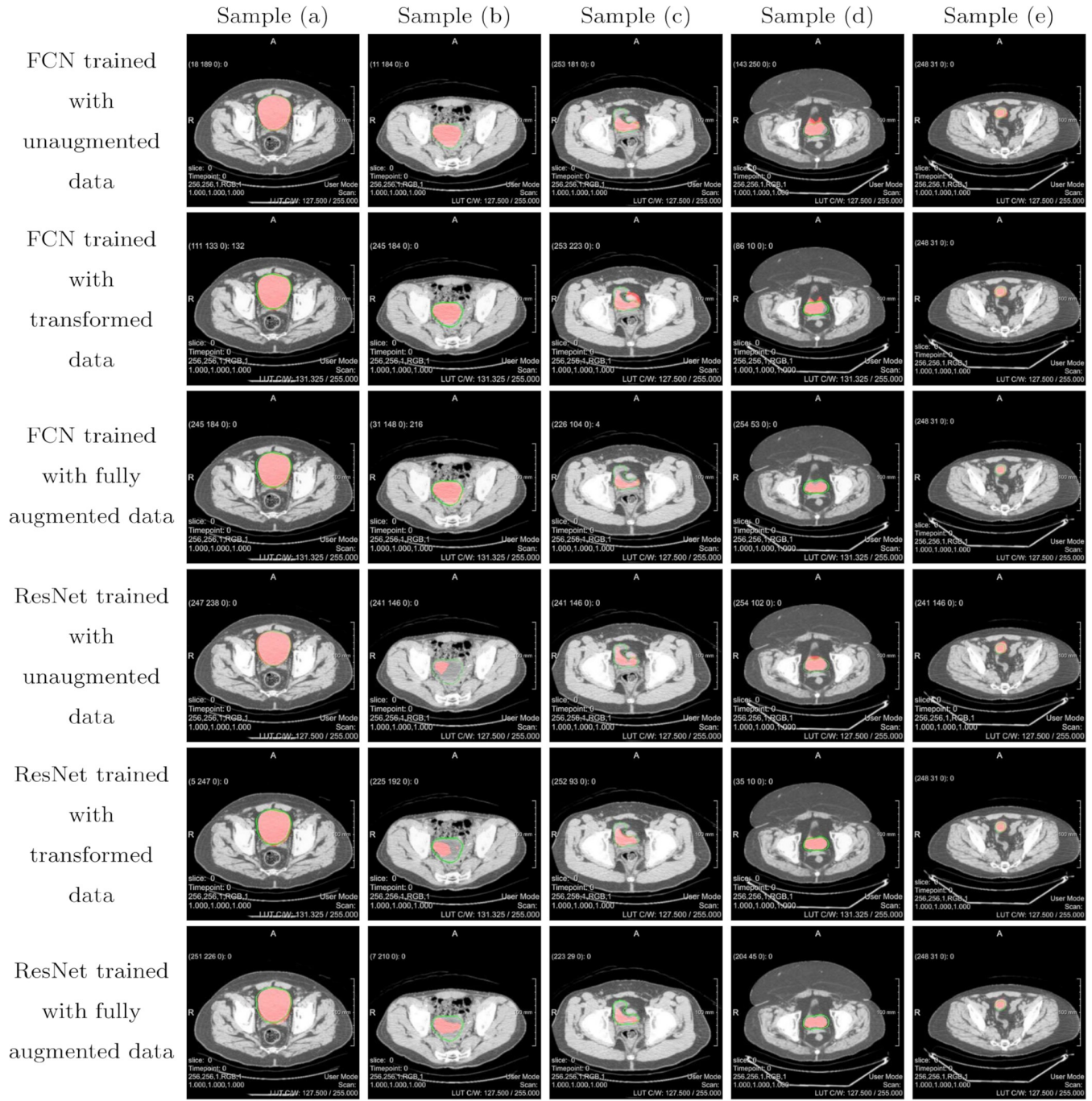

|  | Sample (a) | Sample (b) | Sample (c) | Sample (d) | Sample (e) |
|---|---|---|---|---|---|
| FCN trained with unaugmented data | | | | | |
| FCN trained with transformed data | | | | | |
| FCN trained with fully augmented data | | | | | |
| ResNet trained with unaugmented data | | | | | |
| ResNet trained with transformed data | | | | | |
| ResNet trained with fully augmented data | | | | | |

**Fig 10. Qualitative segmentation result overlays for images scaled to 256x256.** Ground truth labels are shown by the contours in green; the predictions made by the deep learning models are overlaid in red.

Figs 10 and 11 show several representative examples of the obtained segmentation results for images downsampled to a resolution of 256x256 and images at original resolution of 512x512, respectively. For better illustration, original image data was overlaid with the contour of the ground truth in green as well as the prediction made by our deep networks in red. The qualitative segmentation results for images scaled to 256x256 illustrate that for input images with good soft tissue contrast, a large, homogeneous area surface and a regular shape of the urinary bladder, as seen in sample (a), all network models perform well. In sample (b), contrast between the urinary bladder and surrounding tissue is not ideal, moreover, the bladder itself includes varying grey values, meaning that the area surface is not homogeneous. It is evident that while our FCN architecture has no trouble in detecting the urinary bladder in these images, the upsampled ResNet architecture performs poorly. Apparently the upsampled ResNet models are more sensitive against contrast and grey values. However, it can be seen from sample (c), that the ResNet models are better in adapting to distinct shapes. Sample (d) is interesting because our generated ground truth annotation does not follow the very unusual shape of the bladder very well in this example. While the ResNet models seem to fit better to the ground truth label, the segmentation predicted by the FCN models, especially the network trained with transformed data, seems to correspond better to the actual outline of the bladder. Sample (e) shows, that despite our initial concerns, the proposed models are able to identifying the urinary bladder when only a small portion of it is visible in a slice, as long as contrast is good. In fact, the predictions made by our models in some cases even follow the outline of the urinary bladder better than our underlying ground truth segmentation. The same observations can be made when looking at the qualitative segmentation results for images of higher resolution in Fig 11. For input images of high quality, both architectures perform well. FCN does better when segmenting images with low contrast and inhomogeneous grey values as seen in sample (b), while ResNet adapts better to unusual shapes as in sample (c). Again, sample (d) allows for some very interesting observations. Here, our upsampled ResNet architecture does a very good job in detecting the urinary bladder, even recognising the small, detached portion of the bladder at the top. The FCN architecture also produces a more accurate segmentation than our underlying ground truth, but in this case, upsampled ResNet trained with augmented data performs very well. It is also notable that qualitative results for upsampled ResNet trained with unaugmented data of images with resolution 512x512 are worst amongst all achieved predictions. Segmentation results appear very uneven and edged, also they show a lot of outliers which is supported by the high Hausdorff distance of averagely 23.9 pixels for this model. Obviously, a network for higher resolution images also has more parameters that need to be tuned, and in this case, the 630 unaugmented training images apparently did not provide sufficient information to specify all these parameters correctly. It can be noted that while our networks trained with images of higher resolution don't necessarily show a better segmentation performance in terms of our evaluation metrics, as seen in Tables 2 and 3, qualitative results are to some extend better for images with resolution 512x512, especially for network models trained with augmented data. The reason for this is our non-perfect ground truth. In many cases, predictions made by our models don't fit the ground truth we compare it to accurately, which results in low evaluation scores. Nevertheless, looking at the image data one can see that the predictions correspond well with the actual outline of the urinary bladder.

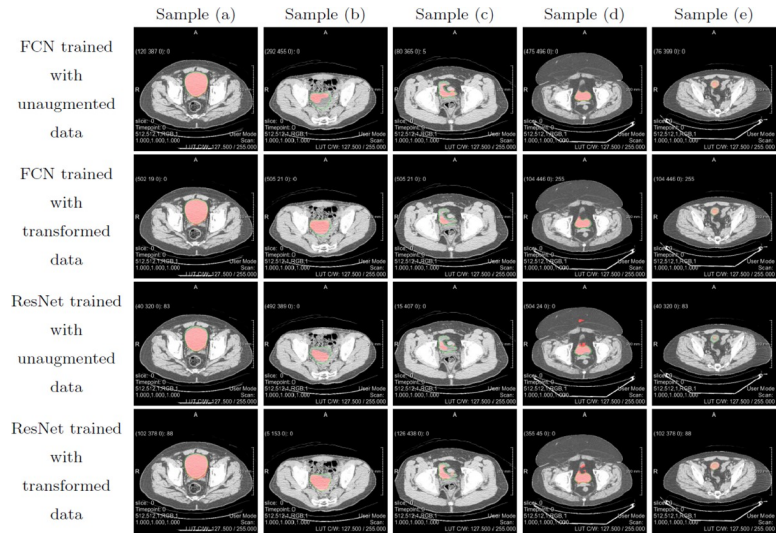

**Fig 11. Qualitative segmentation result overlays for images with resolution 512x512.** Ground truth labels are shown by the contours in green; the predictions made by the deep learning models are overlaid in red.

## Discussion

Small dataset sizes and the lack of annotations due to the complexity of manual segmentation are big limitations to deep learning applications in medical image processing. There have been some attempts to overcome this obstacle, including the usage of existing tools to create labels for pre-training [26], the usage of sparse annotations [27] or the generation of artificial data with Generative Adversarial Networks (GANs) [28]. We introduce a new solution to this problem by generating training and testing datasets for deep learning algorithms by exploiting $^{18}$F-FDG accumulation in the urinary bladder to produce ground truth labels, and by the application of data augmentation to enlarge a small dataset. We showed that when making use of combined PET/CT data, an automatic low-level segmentation of PET image data can be used to attain a fully automatic, high-level segmentation of corresponding CT data. We achieved satisfying segmentation results with a comparably very small image database and completely without the usage of manually segmented image data. Since combined scanners are becoming increasingly more widespread, it can be expected that more, larger PET/CT image databases will be available in the future. Our approach presents a promising tool for automatically processing such databases and can be generalized to all applications of combined PET/CT or combined PET/MRI, such as cancerous tumours in the lung or in the head and neck area, just to name a few.

We used the generated data to train and test two different well-known deep learning models for semantic image segmentation. Our qualitative results show that the proposed segmentation methods can accurately segment the urinary bladder in CT images and are in many cases more accurate than the ground truth labels obtained from PET image data. It is shown that the used FCN architecture generally performs better in terms of evaluation metrics than the proposed ResNet architecture. We achieved the best segmentation performance with our FCN network which was trained with transformed image data. Future work would include a more sophisticated post-processing. In many publications, including in [14] by Chen et al., fully-connected conditional random fields are used to accurately recover object boundaries that are smoothed within the deep neural network. In our case, this might especially improve performance in cases were the urinary bladder has irregular, distinct shapes.

We demonstrated that training data augmentation in the form of transformations, like rotation and scaling, can significantly improve the performance of segmentation networks, however, the addition of zero-mean Gaussian noise to the training data did not result in an enhanced performance in our case. Subsequent work could go into further exploring the effects of data augmentation on the segmentation results, by generating even bigger augmented datasets and by applying different noise types to the original image data.

## Acknowledgments

This work received funding from the Austrian Science Fund (FWF) KLI 678-B31: "enFaced: Virtual and Augmented Reality Training and Navigation Module for 3D-Printed Facial Defect

Reconstructions" (PIs: Jürgen Wallner and Jan Egger) and the TU Graz Lead Project (Mechanics, Modeling and Simulation of Aortic Dissection). Moreover, this work was supported by CAMed (COMET K-Project 871132) which is funded by the Austrian Federal Ministry of Transport, Innovation and Technology (BMVIT) and the Austrian Federal Ministry for Digital and Economic Affairs (BMDW) and the Styrian Business Promotion Agency (SFG).

## Author Contributions

**Conceptualization:** Christina Gsaxner, Jan Egger.

**Data curation:** Jürgen Wallner.

**Formal analysis:** Christina Gsaxner, Jan Egger.

**Funding acquisition:** Jürgen Wallner, Jan Egger.

**Investigation:** Jan Egger.

**Methodology:** Christina Gsaxner, Jan Egger.

**Project administration:** Jürgen Wallner, Jan Egger.

**Resources:** Peter M. Roth, Jürgen Wallner, Jan Egger.

**Software:** Christina Gsaxner.

**Supervision:** Jürgen Wallner, Jan Egger.

**Validation:** Christina Gsaxner, Jan Egger.

**Visualization:** Christina Gsaxner.

**Writing – original draft:** Jan Egger.

**Writing – review & editing:** Christina Gsaxner, Jan Egger.